\documentclass[10pt,twocolumn,letterpaper]{article}

\usepackage[pagenumbers]{cvpr} 

\usepackage{multirow}

\usepackage{placeins}
\usepackage{tikz}
\usepackage{fontawesome5}
\usetikzlibrary{shadows.blur}
\definecolor{cvprblue}{rgb}{0.21,0.49,0.74}
\definecolor{abstractbg}{RGB}{242,247,255}
\usepackage[pagebackref,breaklinks,colorlinks,allcolors=cvprblue]{hyperref}

\def\confName{CVPR}
\def\confYear{2027}

\title{Future Prediction as Training-Only Video Supervision

for Pretrained VLA Policies}

\author{
\textbf{Hanwen Cui}\textsuperscript{1,2,3,*} \quad
\textbf{Yuanyuan Li}\textsuperscript{1,2,*} \quad
\textbf{Qing Guo}\textsuperscript{4} \quad
\textbf{Zhida Zhang}\textsuperscript{3} \\
\textbf{Chao Chen}\textsuperscript{3,\textdagger} \quad
\textbf{Xi Chen}\textsuperscript{3} \quad
\textbf{Jie Cao}\textsuperscript{1,2,\textdaggerdbl} \\[0.5ex]
{
\small\textsuperscript{1}MAIS \& NLPR, Institute of Automation, Chinese Academy of Sciences
} \\
{
\small\textsuperscript{2}School of Artificial Intelligence, University of Chinese Academy of Sciences
} \\
{
\small\textsuperscript{3}Gengine
} \\
{
\small\textsuperscript{4}School of Computer Science and Engineering, Central South University
} \\[0.8ex]
{\small
\textsuperscript{*}Equal contribution \quad
\textsuperscript{\textdagger}Project lead \quad
\textsuperscript{\textdaggerdbl}Corresponding author
}
}

\begin{document}
\raggedbottom 

\makeatletter
\twocolumn[{
\begin{@twocolumnfalse}
  \vspace*{0.15in}
  \noindent
  \includegraphics[height=0.42in]{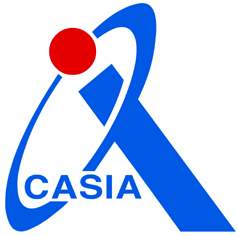}
  \hspace{0.10in}
  \includegraphics[height=0.42in]{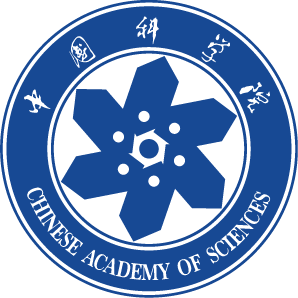}
  \hspace{0.10in}
  \includegraphics[height=0.42in]{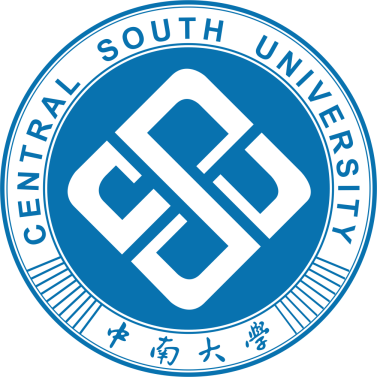}
  \hfill
  \includegraphics[height=0.37in]{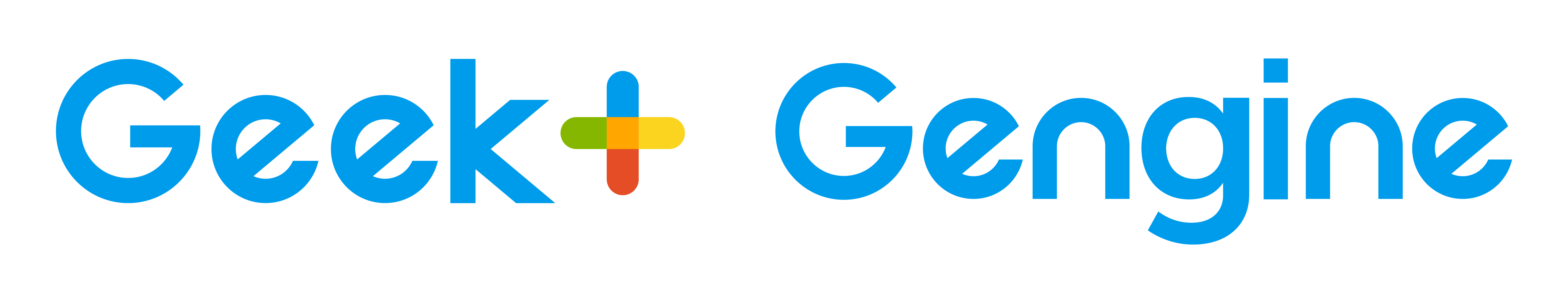}
  \par\vspace{0.35em}

  \begin{center}
    {\LARGE\bfseries \@title\par}
    \vspace{1.15em}
    {\large
      \lineskip 0.5em
      \begin{tabular}[t]{c}
        \@author
      \end{tabular}\par
    }
  \end{center}
  \vspace{0.8em}

  \begingroup
  \renewenvironment{abstract}{%
    \noindent\makebox[\linewidth][c]{\large\bfseries Abstract}\par
    \vspace{0.35em}\normalfont\small\noindent\ignorespaces
  }{\par}
  \noindent\hfill
  \begin{tikzpicture}
      \node[
        fill=abstractbg,
        rounded corners=8pt,
        inner sep=11pt,
        outer sep=0pt,
        blur shadow={shadow blur steps=8,shadow blur radius=0.7ex,shadow opacity=24,shadow xshift=3pt,shadow yshift=-3.5pt}
      ] {%
        \begin{minipage}{\dimexpr\textwidth-22pt\relax}
%
%
%
%

\begin{abstract}

World action models (WAMs) introduce predictive supervision into robot policy learning by jointly modeling future visual observations and actions. Pretrained vision-language-action (VLA) policies acquire rich visual and semantic representations as well as action-generation capabilities. Their action objectives typically lack direct supervision for predicting future visual observations. Future videos in robot demonstrations record object motion and task progress. They can provide temporal visual supervision that complements action labels.

We propose Training-Only Video Supervision, which introduces future-video prediction as an auxiliary objective for VLA fine-tuning. We condition a randomly initialized video-prediction expert on the VLA's shared vision-language features. The expert learns to predict future video latents from robot demonstrations. The video prediction loss and the original action loss jointly update the shared vision-language backbone, providing additional temporal visual supervision. After training, we remove the video expert and its adapter. The fine-tuned VLA retains its original action-generation pathway at inference.
On the 50-task RoboTwin 2.0 benchmark, our method achieves 71.18\% average success, compared with $\pi_{0.5}$ of 58.33\%. In a matched evaluation on four real-world manipulation tasks, success increases from 42.50\% to 61.25\%. These results show that future visual modeling can serve as auxiliary training supervision for pretrained VLAs. The fine-tuned action policy benefits from this supervision without retaining a video-prediction branch at deployment.

\end{abstract}

          \vspace{0.55em}

          \noindent\small
          \faHome\enspace\textbf{Project Page:}\enspace
          \href{https://training-only-video-supervision.pages.dev/}{%
            \textcolor{cvprblue}{\texttt{https://training-only-video-supervision.pages.dev/}}%
          }
        \end{minipage}%
      };
  \end{tikzpicture}
  \hfill\mbox{}\par
  \endgroup

  \vspace{0.35em}
\end{@twocolumnfalse}
}]
\makeatother

\section{Introduction}
\label{sec:introduction}

\begin{figure*}[t]
    \centering
    \includegraphics[width=\textwidth]{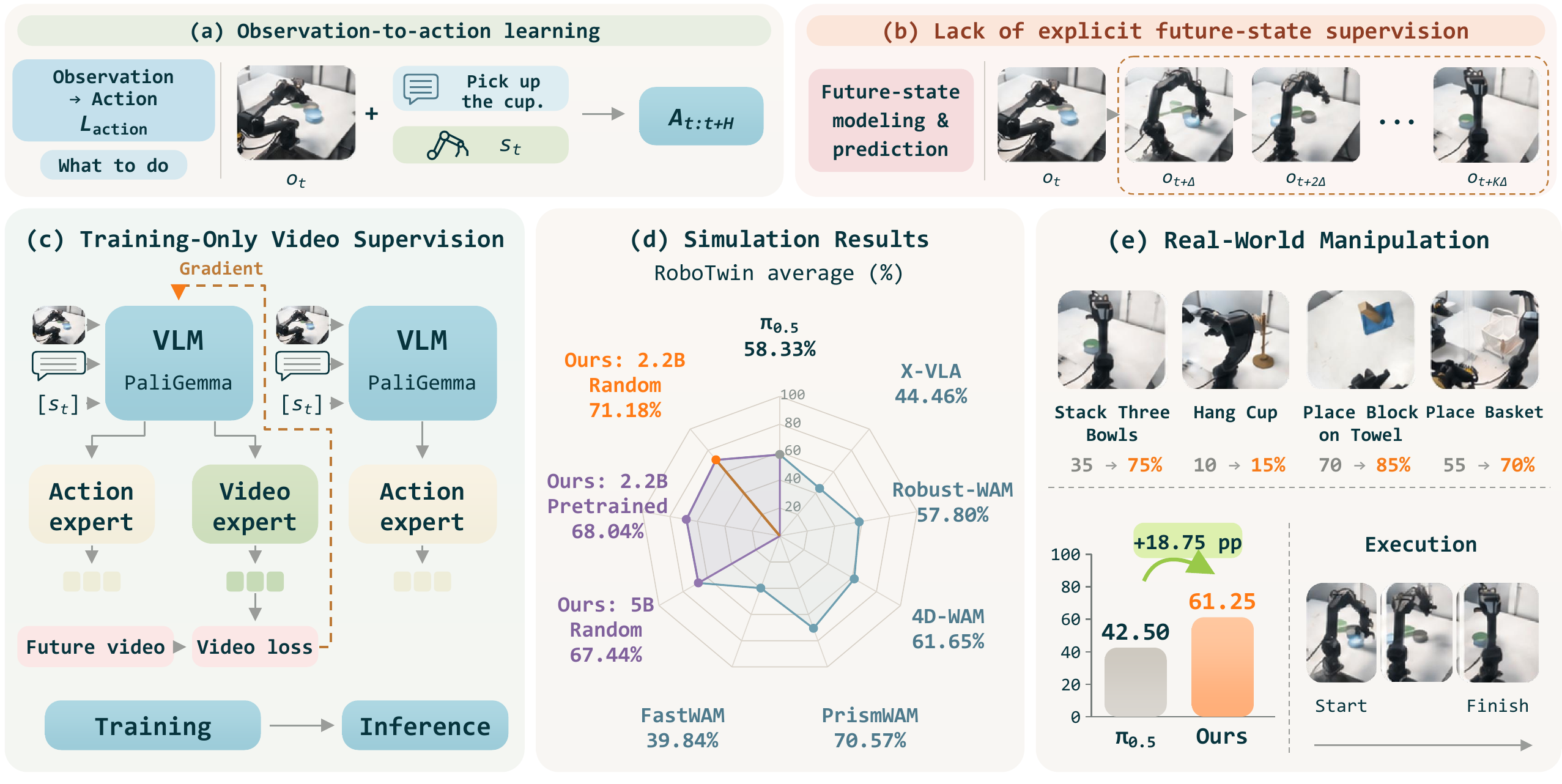}
    \caption{\textbf{Future-Video Prediction as Training-Only Supervision for Pretrained VLA Policies.}
    \textit{(a,b):} Pretrained VLAs learn to map current visual observations to actions conditioned on language instructions and proprioceptive state. However, their original action-learning objective primarily constrains the correspondence between current observations and target actions, providing limited direct supervision for how the environment changes visually after action execution. The dashed box highlights future observations in the demonstrations, which can provide additional temporal visual supervision.
    \textit{(c):} Our training-time video-prediction objective introduces additional temporal visual supervision into the shared representations of the pretrained VLA. The video-prediction loss updates the shared VLM through auxiliary gradients (orange dashed arrow), while gray arrows denote forward computation. The video expert and its adapter are removed at inference.
    \textit{(d):} Reported average success rates on RoboTwin 2.0 for the original $\pi_{0.5}$ reference, recent WAM methods, and our variants. Randomly initialized DiT-2.2B achieves 71.18\%, outperforming the original reference (58.33\%) and the pretrained DiT-2.2B variant (68.04\%), indicating that performance gains can be achieved without pretrained Video DiT generation weights under our experimental settings.
    \textit{(e):} Four real-world tasks and representative execution frames. Overall success increases from 42.50\% to 61.25\% (+18.75 percentage points), over 80 trials per policy.}
    \label{fig:overview}
\end{figure*}


World action models (WAMs) introduce predictive supervision into robot policy learning by jointly modeling future visual observations and actions~\cite{wu2024gr1,guo2024pad,cai2026internw0}.
By predicting future observations, these models learn from visual changes over time as well as demonstrated actions.
This predictive supervision can also augment the fine-tuning of existing pretrained vision-language-action (VLA) policies.
Such policies acquire visual and semantic representations from vision-language pretraining, along with action-generation capabilities learned from robot demonstrations~\cite{zitkovich2023rt2,kim2024openvla,black2025pi05}.
However, their action imitation objectives do not themselves provide targets for predicting subsequent visual states.
The demonstration videos also record object motion, changes in spatial relationships, and task progress.
Using these future observations as prediction targets can encourage policy representations to capture visual changes relevant to manipulation.
Future-video prediction therefore offers a source of temporal visual supervision that can complement action labels during pretrained VLA fine-tuning.


Recent studies incorporate future visual information into policy learning through representation prediction~\cite{zheng2025flare,sun2026vlajepa} or foresight supervision from a frozen pretrained video generator~\cite{ma2026internvlaa15}.
We study future-video prediction in VAE latent space as an auxiliary flow-matching objective during VLA fine-tuning.
To this end, we propose \textbf{Training-Only Video Supervision}.
We build on pretrained $\pi_{0.5}$ and retain its vision-language backbone and action expert.
In our main implementation, a randomly initialized Video DiT expert is conditioned on shared vision-language features through a trainable adapter.
A frozen VAE encodes robot demonstration videos to provide future video-latent prediction targets.
The video loss and the original action loss jointly update the shared vision-language backbone.
The private action expert continues to be trained only by the action loss.
Future visual targets can thus shape the representations used for action generation without changing the action expert's training objective.
After training, we remove the video expert and its adapter.
The deployed policy uses only the fine-tuned VLA and follows its original action-generation pathway.
Figure~\ref{fig:overview} summarizes this training scheme.


We evaluate the method on RoboTwin 2.0, LIBERO, LIBERO Plus, and four real-world manipulation tasks.
RoboTwin includes 50 trained tasks in clean and randomized scenes, while LIBERO Plus tests policies under diverse perturbations.
We compare our benchmark results with $\pi_{0.5}$ reference scores, using published results for RoboTwin 2.0.
A lightweight Transformer video predictor is evaluated on the same RoboTwin tasks under the same training and evaluation protocol as the Video DiT variants.
We further compare video experts with different architectures and initialization schemes to examine how predictor configurations affect policy performance.


On the 50-task RoboTwin 2.0 benchmark, our randomly initialized 2.2B variant achieves 71.18\% average success, compared with the published $\pi_{0.5}$ reference of 58.33\%.
In a matched evaluation on four real-world tasks, success increases from 42.50\% to 61.25\%.
The policy also achieves 98.05\% on LIBERO and 85.15\% on LIBERO Plus.
The lightweight predictor attains slightly higher clean success but lower randomized and average success than the 2.2B variant.
Under the tested training budget, random 2.2B has the highest observed average RoboTwin success among our expert configurations.
These results support future-video prediction as an auxiliary training objective for pretrained VLAs and show that predictor configurations affect performance across evaluation settings.



Our contributions are as follows:
\begin{itemize}
    \item We propose Training-Only Video Supervision, which uses future-video prediction as an auxiliary fine-tuning objective for pretrained VLAs through shared vision-language features.
    The video branch is removed after training, retaining the original action-generation pathway.

    \item We demonstrate competitive performance on RoboTwin 2.0, LIBERO, and LIBERO Plus, and improved success on four real-world tasks under matched training and evaluation conditions.

    \item We show that pretrained Video DiT weights are not required to obtain benefits from future-video supervision.
    Under our data and training budget, the randomly initialized 2.2B expert achieves higher observed success than its pretrained counterpart.
\end{itemize}
\section{Related Work}
\label{sec:related_work}

\begin{figure*}[t]
    \centering
    \includegraphics[width=\textwidth]{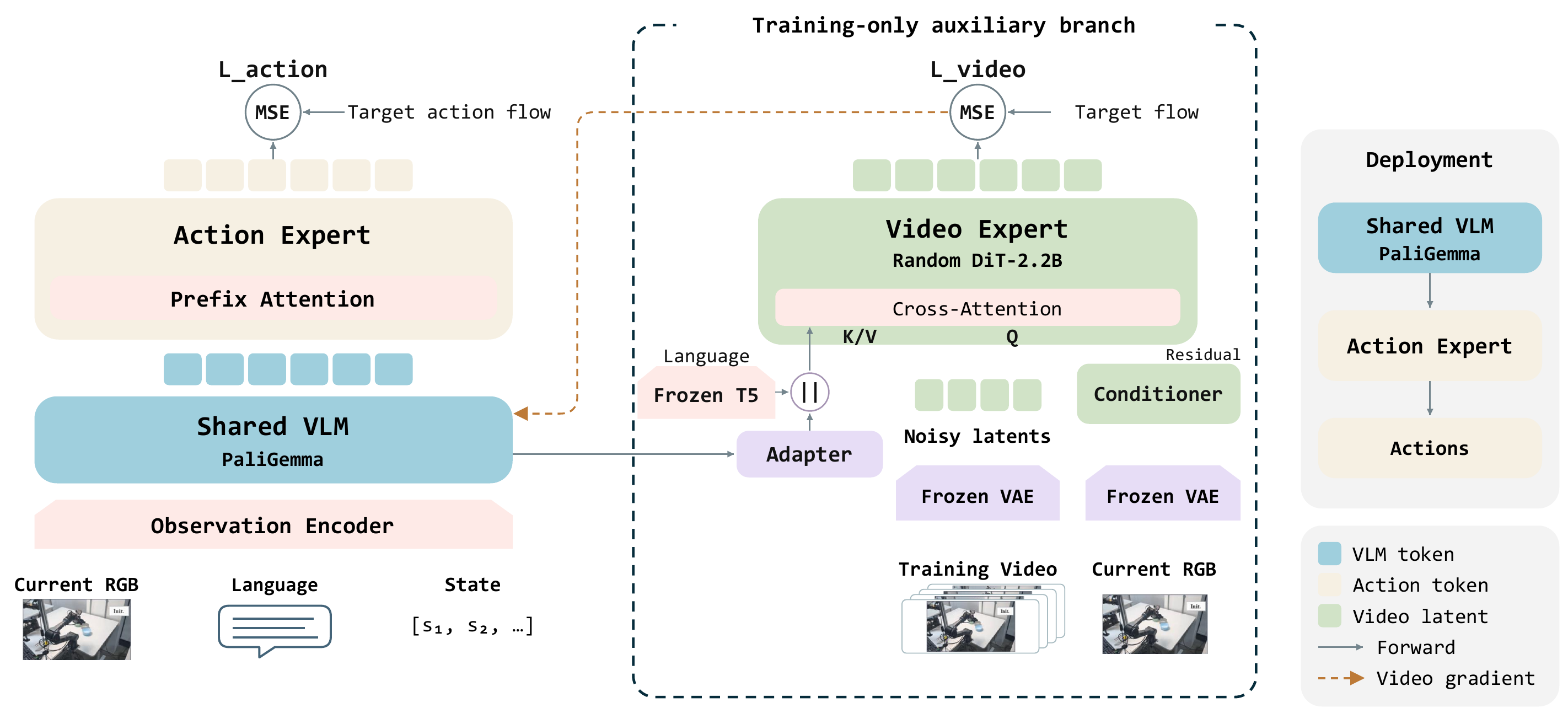}
    \caption{\textbf{Training-only video supervision for $\pi_{0.5}$.} The shared VLM processes current observations, language, and state to condition the action expert and a training-only video expert. A frozen video encoder supplies video latents, and the randomly initialized DiT-2.2B expert learns future-latent flow prediction conditioned on shared VLM features. The dashed orange arrow shows the video-loss gradient updating the shared VLM. The auxiliary branch is removed at inference, leaving only the shared VLM and action expert for deployment.}
    \label{fig:method_arch}
    \label{fig:pi05_auxiliary}
\end{figure*}

\paragraph{Vision-Language-Action Models.}
Vision-Language-Action (VLA) models combine large-scale vision-language pretraining with robot demonstrations to learn policies that generate actions from visual observations and natural-language instructions.
RT-2 represents robot actions as text tokens, transferring vision-language knowledge to robotic control~\cite{zitkovich2023rt2}, while Open X-Embodiment, Octo, OpenVLA, and X-VLA advance generalist VLA learning through large-scale heterogeneous robot datasets, open generalist policies, and cross-embodiment architectures~\cite{openx2023,team2024octo,kim2024openvla,zheng2026x}.
For action generation, Diffusion Policy, $\pi_0$, and $\pi_{0.5}$ explore diffusion-based action prediction, Flow Matching for continuous action generation, and policy learning with heterogeneous co-training and high-level semantic prediction, respectively~\cite{chi2025diffusion,black2024pi_0,black2025pi05}.
Recent studies further explore the use of future visual information to assist VLA learning.
CoT-VLA and $\pi_{0.7}$ use future images or visual subgoals as additional conditions for action generation; FLARE and VLA-JEPA learn about environmental changes by predicting future visual representations in latent space; and InternVLA-A1.5 uses a frozen pretrained video-generation model to supervise foresight tokens, removing the video branch during inference~\cite{zhao2025cotvla,intelligence2026pi,zheng2025flare,sun2026vlajepa,ma2026internvlaa15}.
In contrast, our work introduces explicit future-video prediction as an independent training-time auxiliary objective, providing additional temporal supervision to an existing pretrained VLA through its shared vision-language representations without relying on pretrained video-generation weights.

\paragraph{World Action Models.}
Early studies explored video-generation models for robotic planning and control, including video-based task planning and behavior prediction approaches such as UniPi, Video Language Planning, RoboDreamer, and Dreamitate~\cite{du2023learning,ICLR2024_84878437,zhou2024robodreamer,liang2024dreamitate}.
Building on these efforts, World Action Models (WAMs) further combine future visual prediction with robot action learning.
GR-1, GR-2, PAD, and InternW0 jointly model future observations and robot actions through video-generation pretraining, joint denoising, and collaborative video-action expert architectures; DreamZero, Motus, and the LingBot-VA series further explore video diffusion, autoregressive prediction, and unified video-action pretraining~\cite{wu2024gr1,cheang2024gr,guo2024pad,cai2026internw0,ye2026world,bi2026motus,li2026causal,zhang2026native}.
Recent studies investigate further design choices in WAMs.
FastWAM distinguishes video co-training during training from future-video generation during inference, while retaining a pretrained Video DiT as a single-pass world encoder at deployment~\cite{yuan2026fastwam}.
OpenWAM systematically studies generative backbones, visual representations, and world-action interactions, while ImageWAM, Robust-WAM, 4D-WAM, and PrismWAM explore image-editing priors, semantic foresight, spatiotemporal trajectory representations, and decoupled state representations, respectively~\cite{wang2026openwam,zhang2026imagewam,yan2026robust,yang20264d,xin2026prismwam}.
Unlike prior work that primarily focuses on joint video-action modeling and world-representation design, our work introduces a randomly initialized video-prediction expert as an auxiliary training branch for an existing VLA.
The expert provides additional supervision through future-video prediction and is removed after training, eliminating the need to retain a video-generation model at deployment.

\section{Method}
\label{sec:method}

We propose a training-only video supervision method for pretrained vision-language-action (VLA) policies. We investigate whether future-video prediction can provide useful auxiliary supervision to an existing VLA without relying on pretrained video-generation weights. To this end, we attach a randomly initialized video-prediction expert to the pretrained $\pi_{0.5}$ policy, leaving its original action-generation pathway intact. During training, the action and video-prediction objectives jointly optimize the shared vision-language prefix representations, allowing future visual observations to provide additional supervision for the policy. At inference, the video branch is discarded and the original action pathway is retained. Thus, future-video prediction serves as a training-time learning signal rather than an intermediate prediction required for action generation. Figure~\ref{fig:method_arch} illustrates the overall architecture.

\subsection{Task Definition and Shared Policy Representations}
\label{sec:task_and_backbone}

Given current multi-view visual observations $o_t$, a language instruction $\ell$, and the robot proprioceptive state $s_t$, the policy predicts a chunk of $H$ continuous actions, $A_t=(a_t,\ldots,a_{t+H-1})$. Meanwhile, the training demonstrations also contain subsequent visual observations corresponding to action execution, providing additional supervision for future-video prediction.

We adopt the pretrained VLA $\pi_{0.5}$~\cite{black2025pi05} as the policy backbone. Its shared vision-language prefix consists of a SigLIP visual encoder and a PaliGemma backbone. This module jointly encodes the current multi-view observations, language instruction, and robot state to obtain a shared policy representation:
\begin{equation}
    h_t = F_\theta(o_t,\,\ell,\,s_t),
    \label{eq:prefix}
\end{equation}
where $F_\theta$ denotes the shared vision-language encoding process. The representation $h_t$ retains sequence-level information about the current scene, task semantics, and robot state, and serves as the conditioning input to the original action expert.

In the original $\pi_{0.5}$, the action expert conditions on the shared representation $h_t$ and follows the original Flow Matching mechanism to predict continuous actions. Building on this structure, we connect future-video prediction to the same shared representation, allowing predictive supervision to directly shape the vision-language representations used by the original action policy, rather than merely training an independent video predictor.


%
%
%
%
%
%
%
%
%
%
%


\subsection{Auxiliary Future-Video Prediction}
\label{sec:video_prediction}

To incorporate future visual information into the representation learning of a pretrained VLA, we attach a video-prediction expert that predicts future video latents from robot demonstrations, conditioned on the shared policy representation and current visual observations. This provides additional temporal supervision to the original policy.

\paragraph{Prediction target.}
We extract multi-view observations spanning the current and future timesteps from the training demonstrations and encode them into video latents using a frozen Wan VAE~\cite{wan2025}:
\begin{equation}
    z = [z_0, z^+],
\end{equation}
where $z_0$ denotes the temporal block corresponding to the current observation, and $z^+$ denotes the blocks corresponding to future observations. The video-prediction expert processes the complete latent sequence, but the prediction loss is applied only to the future blocks. The current block participates in the prediction process without serving as a supervision target.

\paragraph{Video prediction expert and conditioning.}
The video-prediction expert is a Video DiT based on the VACE architecture~\cite{jiang2025vace}. Unlike approaches that rely on pretrained video-generation models, we randomly initialize the video expert and train it from scratch on future-video prediction. This allows us to investigate whether predictive supervision can improve VLA representation learning without pretrained Video DiT generation weights. The Wan VAE and T5 encoder used to extract video latents and text features remain frozen.

To incorporate task-relevant information from the original policy, we use a trainable adapter to map the shared policy representation $h_t$ into the conditioning space of the video expert and fuse it with text features to form the policy condition $c_t$. In addition, the current multi-view image is encoded by the frozen VAE and the VACE condition network to produce a visual condition $q_t$. The two conditions play complementary roles: $c_t$ provides the task-aware representations learned by the VLA, while $q_t$ preserves direct visual information about the current scene. Since the policy-conditioning pathway remains differentiable, the video-prediction loss can backpropagate through it to the shared prefix $h_t$.

Given these conditions, the Video DiT predicts the flow of the current and future video latents, while the prediction error on future temporal blocks provides the auxiliary supervision. This design allows the video expert to use current-scene information to model subsequent visual changes and to introduce future visual supervision into the original VLA through the shared policy representation.


%
%
%
%
%
%
%
%
%
%
%
%
%
%
%
%
%


\subsection{Joint Training Objective and Inference}
\label{sec:training_and_inference}

\paragraph{Training objective.}
We jointly optimize action prediction and future-video prediction. For action prediction, we follow the original Flow Matching objective of the pretrained $\pi_{0.5}$ policy. Conditioned on the noisy actions and the shared policy representation $h_t$, the action expert predicts the corresponding action flow velocity, with the loss defined as:
\begin{equation}
    \mathcal{L}_{\mathrm{action}}
    =
    \mathbb{E}
    \left[
        \left\|
            f_\psi(a_{\sigma_a},\sigma_a\mid h_t)
            -(\epsilon_a-a)
        \right\|_2^2
    \right],
    \label{eq:action_loss}
\end{equation}

For future-video prediction, the video expert predicts the corresponding flow velocity from the noised video latent sequence and the conditioning features $c_t$ and $q_t$. Although the model processes the complete sequence containing current and future observations, the prediction error is computed only over valid future temporal blocks:
\begin{equation}
    \mathcal{L}_{\mathrm{video}}
    =
    \mathbb{E}
    \left[
        \operatorname{MSE}_{\mathcal M^+}
        \left(
            \hat v_v^+,\epsilon_v^+-z^+
        \right)
    \right],
    \label{eq:video_loss}
\end{equation}
where $\mathcal M^+$ denotes the mask of valid future temporal blocks. The current temporal block participates in the prediction process but is not used as a supervision target.

\paragraph{Joint optimization and parameter updates.}
The overall training objective is:
\begin{equation}
    \mathcal{L}
    =
    \mathcal{L}_{\mathrm{action}}
    +
    \lambda\mathcal{L}_{\mathrm{video}},
    \label{eq:total_loss}
\end{equation}
where $\lambda$ is the weight of the auxiliary video loss.

The two objectives operate through different parameter pathways:
\[
    \mathcal{L}_{\mathrm{action}}
    \longrightarrow
    \{\theta,\psi\},
    \qquad
    \mathcal{L}_{\mathrm{video}}
    \longrightarrow
    \{\theta,\phi,\omega\}.
\]
The shared vision-language prefix parameters $\theta$ receive supervision from both objectives. The action-expert parameters $\psi$ are updated only by the action loss, whereas the video-prediction expert and its adapter, $\{\phi,\omega\}$, are updated only by the video loss. Thus, future-video prediction improves the shared representations used by the original policy through auxiliary gradients, without directly changing the optimization objective of the action expert.

\paragraph{Inference.}
After training, we remove the video-prediction expert and its adapter, retaining only the updated shared vision-language prefix and the original action-generation expert. The resulting policy follows the original $\pi_{0.5}$ action-generation pathway without additional video-prediction computation.


\section{Experiments}
\label{sec:experiments}

We investigate whether future-video prediction can improve pretrained $\pi_{0.5}$ as an auxiliary training objective.
Our experiments compare a lightweight Transformer with video DiT experts at two scales, examining predictor configuration and initialization.
Simulation evaluations cover RoboTwin 2.0, LIBERO, and LIBERO Plus.
Real-world experiments compare the base policy with our final randomly initialized DiT-2.2B variant on four manipulation tasks.

\begin{table*}[t]
    \centering
    \caption{Simulation success rates (\%). Random and Pretrained denote initialization of the auxiliary video-prediction expert. Bold indicates column maxima. The $\pi_{0.5}$ scores on RoboTwin and standard LIBERO are official references; its Plus score is our evaluation.}
    \label{tab:simulation_comparison}
    \small
    \setlength{\tabcolsep}{4pt}
    \begin{tabular*}{\textwidth}{@{\extracolsep{\fill}}lccccc@{}}
        \toprule
        & \multicolumn{3}{c}{RoboTwin 2.0} & \multicolumn{2}{c}{LIBERO} \\
        \cmidrule(lr){2-4}\cmidrule(lr){5-6}
        Method & Clean$\rightarrow$Clean & Clean$\rightarrow$Random & Average & Standard & Plus \\
        \midrule
        FastWAM & 77.76 & 1.92 & 39.84 & 97.60 & 52.56 \\
        X-VLA & 67.96 & 20.95 & 44.46 & 98.10 & 71.40 \\
        Robust-WAM & 81.20 & 34.40 & 57.80 & 97.90 & 58.90 \\
        4D-WAM & 81.50 & 41.80 & 61.65 & 98.60 & 71.01 \\
        PrismWAM & \textbf{87.98} & 53.16 & 70.57 & \textbf{99.35} & 79.44 \\
        \midrule
        $\pi_{0.5}$ & 70.66 & 46.00 & 58.33 & 96.85 & 84.66 \\
        \textbf{$\pi_{0.5}$ + DiT-2.2B (Random)} & 84.84 & \textbf{57.52} & \textbf{71.18} & 98.05 & \textbf{85.15} \\
        \bottomrule
    \end{tabular*}
    \vspace{2pt}
    \begin{minipage}{\textwidth}
        \footnotesize
    \end{minipage}
\end{table*}

\begin{table*}[t]
    \centering
    \caption{Success rates (\%) for $\pi_{0.5}$ with different training-only video experts. Random and Pretrained denote video-expert initialization. Bold indicates the highest score among the five configurations in each column.}
    \label{tab:video_expert_configurations}
    \small
    \setlength{\tabcolsep}{4pt}
    \begin{tabular*}{\textwidth}{@{\extracolsep{\fill}}lccccc@{}}
        \toprule
        & \multicolumn{3}{c}{RoboTwin 2.0} & \multicolumn{2}{c}{LIBERO} \\
        \cmidrule(lr){2-4}\cmidrule(lr){5-6}
        Video Expert & Clean & Random & Average & Standard & Plus \\
        \midrule
        Lightweight Transformer (Random)
        & \textbf{85.22} & 53.84 & 69.53 & 96.85 & 84.99 \\
        \midrule
        DiT-5B (Pretrained)
        & 82.32 & 55.70 & 69.01 & 96.95 & \textbf{85.97} \\
        DiT-5B (Random)
        & 83.64 & 51.24 & 67.44 & 97.60 & 85.12 \\
        \midrule
        DiT-2.2B (Pretrained)
        & 81.34 & 54.74 & 68.04 & 97.80 & 84.96 \\
        DiT-2.2B (Random)
        & 84.84 & \textbf{57.52} & \textbf{71.18}
        & \textbf{98.05} & 85.15 \\
        \bottomrule
    \end{tabular*}
\end{table*}

\subsection{Experimental Setup}
\label{sec:experimental_setup}

\paragraph{Models.}
We equip $\pi_{0.5}$ with one of three training-only video-prediction experts: a lightweight Transformer, DiT-2.2B, or DiT-5B.
The lightweight predictor has no corresponding video-generation pretrained checkpoint and is randomly initialized.
For each DiT architecture, we evaluate both pretrained and random initialization, yielding five configurations.
All policies start from pretrained $\pi_{0.5}$ weights; random initialization concerns the auxiliary video expert.

\paragraph{Simulation benchmarks.}
On RoboTwin 2.0~\cite{chen2025robotwin2}, we use the 50-task co-training setting with 50 clean demonstrations per task.
All configurations share fixed evaluation scenes: 100 Clean$\rightarrow$Clean and 100 Clean$\rightarrow$Random episodes per task, totaling 10,000 evaluations per policy.
Randomization changes the scene distribution while retaining the training-task set.

LIBERO~\cite{liu2023libero} comprises the Spatial, Object, Goal, and Long suites.
We evaluate 10 tasks per suite with 50 episodes per task, totaling 2,000 episodes per policy.
LIBERO Plus~\cite{fei2026libero} evaluates standard-LIBERO-trained policies on 10,030 scenarios spanning camera, robot-state, language, lighting, background, sensor-noise, and object-layout perturbations.

\paragraph{Training and inference.}
For RoboTwin, we evaluate action-EMA checkpoints at 50,000 training steps for all five configurations.
Policies predict 32 action steps, execute 24 steps before replanning, and use 10 action-denoising steps.
Each configuration retains its training normalization statistics and action representation.
All video experts are removed at inference, and evaluation uses only the action policy.
Full implementation, training, and benchmark-specific inference settings are provided in the supplementary material.

\paragraph{Baselines.}
We use $\pi_{0.5}$ as the base-policy reference and include X-VLA~\cite{zheng2026x} as a generalist VLA baseline.
We additionally compare against FastWAM~\cite{yuan2026fastwam}, Robust-WAM~\cite{yan2026robust}, PrismWAM~\cite{xin2026prismwam}, and 4D-WAM~\cite{yang20264d}, covering related video-action modeling and predictive-supervision methods.

\paragraph{Metrics.}
We report success rates.
The RoboTwin average equally weights Clean and Random performance; LIBERO averages the four suites; Plus pools successful episodes across all scenarios.
Percentage-point differences are computed before rounding.

\begin{figure*}[t]
    \centering
    \includegraphics[width=\textwidth]{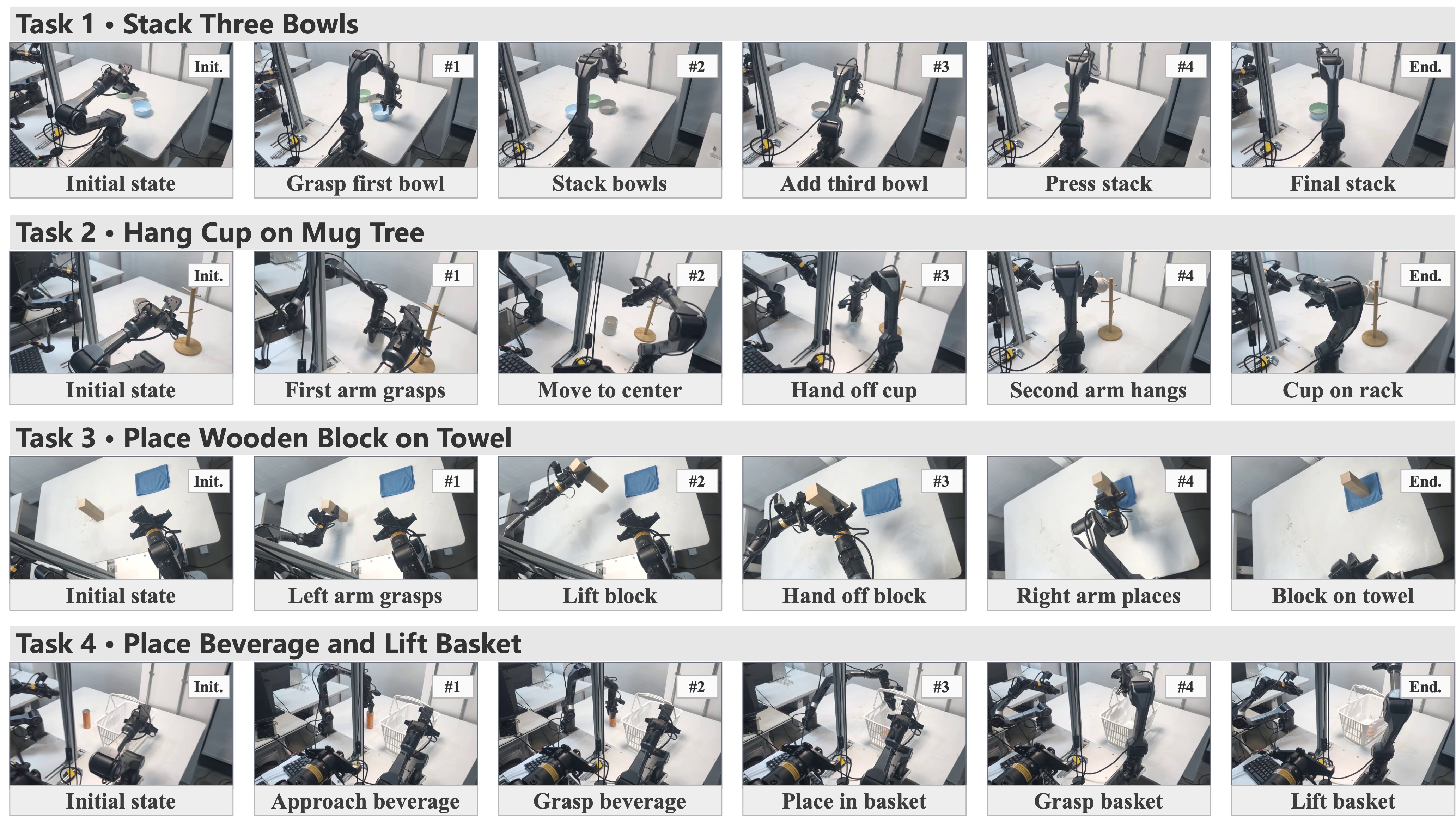}
    \caption{Representative autonomous execution sequences for four real-world manipulation tasks on an AgileX ALOHA bimanual robot platform. We collect 100 teleoperated demonstrations per task (400 in total) and use them to train the policies. After training, the robot autonomously executes each task without human intervention. The four tasks, from top to bottom, are: (1) Stack Three Bowls, which involves grasping and stacking three bowls in sequence; (2) Hang Cup on Mug Tree, which involves bimanual coordination to grasp, hand over, and hang a cup; (3) Place Wooden Block on Towel, which involves bimanual coordination to grasp, hand over, and place a wooden block; and (4) Place Beverage and Lift Basket, which involves placing a beverage in a basket, followed by grasping and lifting the basket. Each row shows six representative execution stages from the initial state to task completion, from left to right.}
    \label{fig:real_robot_sequences}
\end{figure*}

\subsection{Simulation Results}
\label{sec:simulation_results}

Published method comparisons are reported in Table~\ref{tab:simulation_comparison}.
Table~\ref{tab:video_expert_configurations} summarizes the five video-expert configurations used in our analysis.

\subsubsection{Results on RoboTwin 2.0}
\label{sec:robotwin_results}

Random DiT-2.2B achieves 84.84\% Clean success and 57.52\% Random success, yielding the highest average among the methods in Table~\ref{tab:simulation_comparison}, at 71.18\%.
Within our expert configurations, the lightweight predictor has the highest Clean score, whereas random DiT-2.2B achieves the highest Random score and two-domain average.
The final configuration therefore provides stronger aggregate performance without leading every individual condition.

Relative to $\pi_{0.5}$, random DiT-2.2B achieves higher reported success by 14.18 points on Clean, 11.52 points on Random, and 12.85 points on average.
Its average also exceeds those of PrismWAM, Robust-WAM, and 4D-WAM in Table~\ref{tab:simulation_comparison}, indicating competitive aggregate performance across standard and randomized scenes.

\subsubsection{Results on LIBERO and LIBERO Plus}
\label{sec:libero_results}

Random DiT-2.2B achieves the highest standard LIBERO success among our five configurations, at 98.05\%, compared with $\pi_{0.5}$ of 96.85\%.
The lightweight predictor also achieves 96.85\%, while random DiT-2.2B exceeds it by 1.20 percentage points.
This comparison shows higher standard-task success with the 2.2B configuration.

On LIBERO Plus, pretrained DiT-5B achieves the highest score among our configurations, at 85.97\%.
Random DiT-2.2B and random DiT-5B obtain 85.15\% and 85.12\%, respectively, while the lightweight predictor reaches 84.99\%.
Standard LIBERO and Plus therefore favor different expert configurations.
Random DiT-2.2B provides strong performance across benchmarks, but does not attain the highest Plus score.

\subsubsection{Predictive Supervision, Expert Configuration, and Initialization}
\label{sec:prediction_analysis}

\paragraph{Benefits of video-prediction supervision.}
The lightweight predictor provides a starting point for examining auxiliary future-video supervision.
Relative to the base policy, it improves Clean success, the RoboTwin average, LIBERO, and Plus, but reduces Random success.
This pattern indicates that the benefits depend on the predictor configuration and evaluation condition.
Because the lightweight branch does not load pretrained video-DiT generation weights, these observations motivate examining whether a larger randomly initialized video expert can provide more comprehensive benefits.

All reported evaluations remove the video branch at inference.
The resulting performance is therefore obtained by the trained action policy without executing future-video prediction during deployment.
Together with the matched real-world comparison below, these results support future-video prediction as a useful auxiliary training objective for $\pi_{0.5}$.

\paragraph{Predictor architecture and capacity.}
Moving from the lightweight predictor to random DiT-2.2B increases RoboTwin average success from 69.53\% to 71.18\% and Random success from 53.84\% to 57.52\%.
Standard LIBERO success also increases by 1.20 percentage points.
Clean success is slightly lower, by 0.38 points, so the 2.2B configuration improves aggregate performance rather than every individual metric.

Increasing the expert size further to random DiT-5B does not improve success in the evaluated configurations.
Random DiT-2.2B exceeds random DiT-5B by 1.20 points on Clean, 6.28 points on Random, and 3.74 points on the RoboTwin average.
The difference is more pronounced under scene randomization.
Random DiT-2.2B also has higher LIBERO success, while their Plus scores are close.
These results indicate that predictor capacity and control performance do not follow a simple monotonic relationship.

A possible explanation is that the manipulation data and optimization resources are limited relative to video-expert capacity.
With 2,500 RoboTwin demonstrations, the 2.2B expert may be easier to optimize sufficiently than the 5B expert.

\paragraph{Effect of video-expert initialization.}
For DiT-2.2B, random initialization achieves higher success than pretrained initialization across the evaluated benchmarks.
The gains are 3.50 points on Clean, 2.78 points on Random, and 3.14 points on the RoboTwin average, alongside 0.25 points on LIBERO and 0.20 points on Plus.
The RoboTwin advantage is present in both conditions, indicating that its average improvement does not result from a gain in one condition offsetting a decline in the other.
Under these settings, loading video-generation pretrained weights does not further improve action success.

The DiT-5B comparison shows a different pattern.
Pretrained initialization exceeds random initialization by 4.46 points on Random, 1.57 points on the RoboTwin average, and 0.85 points on Plus, but has lower Clean and standard LIBERO success.
The effect of video-generation pretraining therefore depends on the expert configuration and evaluation condition.

Taken together, the lightweight and DiT results show that effective auxiliary video supervision can be obtained without pretrained video-DiT generation weights.
They also show that pretraining can benefit particular configurations and that greater expert capacity does not guarantee higher task success.
The supported conclusion is that video-DiT generation pretraining is not a prerequisite for useful auxiliary supervision, rather than that it is uniformly ineffective.

\subsection{Real-world Results}
\label{sec:real_world_results}

\begin{figure}[t]
    \centering
    \includegraphics[width=\columnwidth]{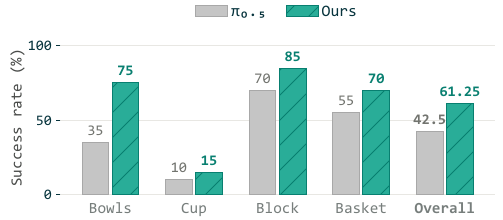}
    \caption{\textbf{Real-world success rates.} Gray: original $\pi_{0.5}$; hatched green: our random DiT-2.2B variant. Each task uses 20 trials per policy; Overall aggregates 80 trials.
    \newline\textit{Task names:} Bowls, Cup, Block, and Basket denote Stack Three Bowls, Hang Cup on Mug Tree, Place Wooden Block on Towel, and Place Beverage and Lift Basket, respectively.}
    \label{fig:real_world_success}
\end{figure}

\paragraph{Setup.}
We compare $\pi_{0.5}$ with $\pi_{0.5}$ + randomly initialized DiT-2.2B on an AgileX ALOHA platform.
The evaluation comprises four manipulation tasks, illustrated in Fig.~\ref{fig:real_robot_sequences}, with per-task success rates in Fig.~\ref{fig:real_world_success}.
Both policies use the same demonstrations, training budget, language instructions, and evaluation conditions.
Camera configurations and control settings are detailed in the supplementary material.

We collect 100 demonstrations per task, totaling 400 demonstrations, and conduct 20 evaluation trials per task and policy, totaling 80 trials per policy.
A trial is successful when the policy completes the task within 30~s without human intervention.
Task-specific success criteria and the timing protocol are provided in the supplementary material.

\paragraph{Results.}
Figure~\ref{fig:real_world_success} reports per-task and overall success rates.
Overall success is the mean across the four tasks.

The random DiT-2.2B auxiliary variant increases overall success from 42.50\% to 61.25\%, adding 15 successful trials across 80 trials per policy and improving success by 18.75 percentage points, as summarized in Fig.~\ref{fig:overview}(e).
Improvements occur on all four tasks, with per-task gains of 40, 5, 15, and 15 points, respectively.
These physical-robot results complement the simulation findings and support auxiliary video prediction for improving action learning.

\section{Conclusion}
\label{sec:conclusion}

We presented Training-Only Video Supervision, which augments pretrained $\pi_{0.5}$ with an auxiliary future-video prediction objective.
Video and action losses jointly optimize shared vision-language representations.
After training, the video expert and its adapter are removed, preserving the original action-generation pathway without additional inference-time video computation.

Our randomly initialized DiT-2.2B configuration achieves 71.18\% average success across the Clean$\rightarrow$Clean and Clean$\rightarrow$Random conditions of the 50-task RoboTwin benchmark, alongside 98.05\% on LIBERO and 85.15\% on LIBERO Plus.
In a matched evaluation on four real-world tasks, overall success increases from 42.50\% to 61.25\%, a gain of 18.75 percentage points.
These results support future-video prediction as a useful auxiliary training objective for pretrained VLA policies.

The lightweight predictor shows that auxiliary prediction can provide benefits without pretrained video-DiT weights.
Scaling to random DiT-2.2B improves RoboTwin average and standard LIBERO performance, while further scaling to random DiT-5B does not yield higher success.
Initialization effects are also configuration-dependent: random DiT-2.2B outperforms its pretrained counterpart, whereas pretrained DiT-5B achieves the highest Plus score.
Under our data and training settings, video-DiT generation pretraining is therefore not a prerequisite for effective auxiliary supervision, and greater capacity does not guarantee better control.
Future work will explore other pretrained VLA architectures and the interactions among data scale, expert design, and initialization.
\section*{LLM Usage Statement}

OpenAI Codex assisted with experiment execution, including launching and coordinating multiple experiments in parallel. The authors are responsible for verifying the experimental configurations, results, and manuscript content, and for all scientific claims.

{
    \small
    \bibliographystyle{ieeenat_fullname}
    \bibliography{main}
}

\clearpage
\maketitlesupplementary
\setcounter{section}{0}
\setcounter{subsection}{0}
\setcounter{subsubsection}{0}
\setcounter{figure}{0}
\setcounter{table}{0}
\setcounter{equation}{0}
\renewcommand{\thesection}{\arabic{section}}
\renewcommand{\thefigure}{S\arabic{figure}}
\renewcommand{\thetable}{S\arabic{table}}
\renewcommand{\theequation}{S\arabic{equation}}
\renewcommand{\theHsection}{supp.\arabic{section}}
\renewcommand{\theHsubsection}{supp.\arabic{section}.\arabic{subsection}}
\renewcommand{\theHsubsubsection}{supp.\arabic{section}.\arabic{subsection}.\arabic{subsubsection}}
\renewcommand{\theHfigure}{supp.\arabic{figure}}
\renewcommand{\theHtable}{supp.\arabic{table}}
\renewcommand{\theHequation}{supp.\arabic{equation}}

\section{Method Implementation and Expert Configurations}
\label{sec:supp_implementation}

We describe the implementation of Training-Only Video Supervision (TOVS), including the auxiliary expert configurations, bimanual observations and conditioning, prediction targets, and gradient paths. The architecture, tensor shapes, and objective implementation below describe the real-world random DiT-2.2B configuration.

\subsection{Policy Backbone and Video Experts}
\label{sec:supp_pi05}
\label{sec:supp_pi05_architecture}

The main paper evaluates five training-time video-expert configurations: a randomly initialized lightweight Transformer, and DiT-2.2B and DiT-5B experts with pretrained or random initialization. All configurations retain pretrained $\pi_{0.5}$ initialization; random initialization refers to the auxiliary video expert.

\paragraph{Random DiT-2.2B architecture.}
The video expert uses the Wan2.1-VACE-1.3B topology, with 30 main Transformer blocks and 15 condition blocks. Both use hidden dimension 1,536, 12 attention heads, and feed-forward dimension 8,960. Condition hints are injected after main blocks $0,2,\ldots,28$.

The expert is randomly initialized with seed 42 without loading pretrained video-generation weights. The Wan2.1 VAE and UMT5-XXL retain their pretrained weights and remain frozen for offline video and task-text encoding. The $\pi_{0.5}$ policy has 18 joint Transformer layers, with shared-prefix width 2,048 and action-expert width 1,024.

The complete video expert contains approximately 2.154B parameters, rounded to DiT-2.2B in the paper. This count includes its condition blocks, whereas the source topology retains the name Wan2.1-VACE-1.3B. The prefix adapter is counted separately in Table~\ref{tab:supp_vace_parameter_counts}.

\subsection{Observations, Targets, and Conditioning}
\label{sec:supp_vace_inputs}

\paragraph{Bimanual observations and actions.}
The real-world bimanual pipeline orders the cameras as top, left wrist, and right wrist. Policy images are resized with preserved aspect ratio and black padding to $224\times224$. Video supervision uses a $384\times320$ composite: a $256\times320$ top-camera image above two $128\times160$ wrist-camera images.
\begin{table}[t]
    \centering
    \caption{Parameter counts of the random DiT-2.2B implementation. The video expert includes main and condition blocks; the training total excludes the frozen offline VAE and T5 encoders.}
    \label{tab:supp_vace_parameter_counts}
    \small
    \begin{tabular*}{\columnwidth}{@{\extracolsep{\fill}}lr@{}}
        \toprule
        Component & Parameters \\
        \midrule
        $\pi_{0.5}$ policy & 3,353,433,872 \\
        Video expert & 2,153,972,032 \\
        Prefix adapter & 8,392,704 \\
        \midrule
        Training total & 5,515,798,608 \\
        \bottomrule
    \end{tabular*}
\end{table}
State and action have 14 physical dimensions: six left-arm joints, the left gripper, six right-arm joints, and the right gripper. Actions specify absolute joint and gripper targets. State and action are normalized using their respective task-specific 1st/99th-percentile statistics and padded to dimension 32. Discretized state and the instruction form a text sequence padded to 200 tokens. Action chunks have shape $[B,32,32]$; their loss covers only the 14 physical dimensions and valid timesteps.

\paragraph{Video targets and offline caching.}
At timestep $t$, video frames are sampled at offsets $[0,4,8,12,16,20,24,28,32]$. The frozen Wan2.1 VAE maps videos of shape $[B,3,9,384,320]$ to latents of shape $[B,16,3,48,40]$, comprising one current block and two future blocks. Beyond an episode endpoint, the final available frame is repeated for encoding, and a two-element validity mask excludes out-of-range future targets. Future-block validity is checked at offsets 16 and 32. The cached features are frozen VAE outputs and T5 features; T5 outputs have shape $[B,128,4096]$ with a token mask.

\paragraph{Policy and text conditioning.}
Each policy image produces 256 visual tokens. Together with 200 instruction/state tokens, the shared prefix contains 968 positions of width 2,048, including masked padding positions. A trainable $2048\!\rightarrow\!4096$ adapter projects these features. The 128 T5 tokens precede the adapted prefix, yielding context of shape $[B,1096,4096]$. The video expert maps this context to width 1,536 and uses it as cross-attention keys and values in both the main and condition blocks. Queries are supplied by the corresponding video or condition tokens. Prefix features are computed online without detachment, allowing video-loss gradients to reach the shared policy representation.

\subsection{Training Objective, Gradients, and Deployment}
\label{sec:supp_vace_objective}

\paragraph{Video flow-matching objective.}
Current and future latent blocks form the complete sequence $z$. Independent Gaussian noise $\epsilon_v$ is applied to all three temporal positions:
\begin{equation}
    z_{\sigma_v}=(1-\sigma_v)z+\sigma_v\epsilon_v,
    \qquad \sigma_v\sim\mathcal{U}(0.001,0.999).
    \label{eq:supp_vace_full_noise}
\end{equation}
The scheduler uses shift 1 and timestep $1000\sigma_v$. The main input's current block is also noised; clean current-scene information is provided by the independent conditioning pathway. Action and video noise are sampled independently.

The expert predicts the full latent flow, but supervision covers only future indices $i=1,2$. For sample $b$, let $\hat v_{bi}$ denote the predicted future-block flow and $m_{bi}$ its validity indicator. The loss is
\begin{equation}
\begin{aligned}
    \ell_{bi}
    &=\operatorname{mean}_{C,H',W'}
      \left[\big(\hat v_{bi}-(\epsilon_v-z)_{bi}\big)^2\right],\\
    \mathcal{L}_{\mathrm{video}}
    &=\frac{\sum_b\sum_{i=1}^{2}m_{bi}\ell_{bi}}
            {\max\!\left(1,\sum_b\sum_{i=1}^{2}m_{bi}\right)}.
\end{aligned}
\label{eq:supp_vace_masked_loss}
\end{equation}

Valid-block counts are aggregated across distributed workers and gradient-accumulation microbatches to prevent padded targets from diluting supervision. Training combines the original action objective with the auxiliary loss:
\begin{equation}
    \mathcal{L}=\mathcal{L}_{\mathrm{action}}
       +\lambda_{\mathrm{video}}\mathcal{L}_{\mathrm{video}}.
    \label{eq:supp_joint_objective}
\end{equation}

\paragraph{Gradient paths and deployment.}
The policy uses a block-triangular prefix/action attention mask. Prefix tokens attend only to valid prefix tokens, whereas action tokens can attend to valid prefix and action tokens. The shared prefix therefore does not read the noisy action suffix. Action loss updates the visual encoder, shared prefix, and private action expert. Video loss updates the shared visual/prefix pathway, adapter, and video expert, without directly updating the private action expert. The VAE and T5 remain frozen.

\begin{table}[t]
    \centering
    \caption{Loss-gradient and deployment roles. Frozen encoders provide cached training features.}
    \label{tab:supp_gradient_roles}
    \small
    \setlength{\tabcolsep}{3pt}
    \begin{tabular*}{\columnwidth}{@{\extracolsep{\fill}}lccc@{}}
        \toprule
        Component & Action & Video & Deployed \\
        \midrule
        Visual encoder / shared VLM & Yes & Yes & Yes \\
        Action expert / head & Yes & No & Yes \\
        Adapter / video expert & No & Yes & No \\
        Frozen VAE / T5 & No & No & No \\
        \bottomrule
    \end{tabular*}
\end{table}

As summarized in Table~\ref{tab:supp_gradient_roles}, deployment executes only the trained visual encoder, shared VLM, and action expert. The video expert and its condition network, prefix adapter, and frozen video/text encoders are removed from the action-generation pathway, introducing no additional video-prediction computation.

\section{Experimental Setup}
\label{sec:supp_training}
\label{sec:supp_current_protocols}

This section describes the simulation and real-world settings for TOVS. The independent FastWAM and OpenWAM protocols are specified in Sec.~\ref{sec:supp_wam_studies}.

\subsection{Simulation Training}

All five auxiliary configurations start from the same pretrained $\pi_{0.5}$ policy weights. RoboTwin training uses 50 clean demonstrations for each of 50 tasks, totaling 2,500 demonstrations. Evaluation uses the 50,000-step action-EMA checkpoints listed in Table~\ref{tab:supp_expert_configurations}. 

\begin{table}[t]
    \centering
    \caption{Auxiliary configurations and RoboTwin evaluation checkpoints. All checkpoints use action EMA.}
    \label{tab:supp_expert_configurations}
    \small
    \setlength{\tabcolsep}{3pt}
    \begin{tabular*}{\columnwidth}{@{\extracolsep{\fill}}llr@{}}
        \toprule
        Video expert & Initialization & Steps \\
        \midrule
        Lightweight Transformer & Random & 50,000 \\
        DiT-5B & Pretrained & 50,000 \\
        DiT-5B & Random & 50,000 \\
        DiT-2.2B & Pretrained & 50,000 \\
        DiT-2.2B & Random & 50,000 \\
        \bottomrule
    \end{tabular*}
\end{table}

\subsection{Simulation Evaluation}

\paragraph{RoboTwin 2.0.}
Each auxiliary configuration is evaluated on the same 100 fixed scenes per task under Clean$\rightarrow$Clean (C2C) and Clean$\rightarrow$Random (C2R), across all 50 tasks. Environmental randomization changes scene conditions while retaining task identities. Each condition contains 5,000 episodes, and the overall score equally weights the two condition rates. Policies predict 32 action steps, execute 24 before replanning, and use ten action-denoising steps.

\paragraph{LIBERO and LIBERO Plus.}
Standard LIBERO comprises the Spatial, Object, Goal, and Long suites, each containing ten tasks evaluated for 50 episodes per task. LIBERO Plus evaluates standard-LIBERO-trained policies on 10,030 scenarios spanning camera viewpoint, robot state, language, lighting, background, sensor noise, and object layout. Standard LIBERO reports the mean of the four suite rates; Plus pools successes over all scenarios. The evaluation budgets and aggregation rules across the three simulation benchmarks are summarized in Table~\ref{tab:supp_eval_budgets}.

\begin{table}[t]
    \centering
    \caption{Simulation evaluation budgets per policy and score aggregation.}
    \label{tab:supp_eval_budgets}
    \small
    \setlength{\tabcolsep}{3pt}
    \begin{tabular*}{\columnwidth}{@{\extracolsep{\fill}}lrl@{}}
        \toprule
        Benchmark & Episodes & Aggregation \\
        \midrule
        RoboTwin 2.0 & 10,000 & Two-condition mean \\
        LIBERO & 2,000 & Four-suite mean \\
        LIBERO Plus & 10,030 & Pooled successes \\
        \bottomrule
    \end{tabular*}
\end{table}

\paragraph{Comparison scope.}
The main paper reports the $\pi_{0.5}$ RoboTwin and standard LIBERO scores as published references, whereas the LIBERO Plus score of 84.66\% is a local evaluation. Auxiliary configurations use the evaluation procedures described here; published references retain their reported settings. Initialization comparisons are interpreted within each DiT architecture, while cross-expert comparisons also change architecture.

\subsection{Real-World Training}

The real-world study uses 100 teleoperated demonstrations for each of four tasks. The base policy and random DiT-2.2B policy use matched demonstrations and training budgets. In the random DiT-2.2B configuration, each task is trained independently with its own instruction and normalization statistics. Cached video and text features are provided by the frozen encoders described in Sec.~\ref{sec:supp_vace_inputs}. The main training settings are summarized in Table~\ref{tab:supp_real_world_training}.

\begin{table}[t]
    \centering
    \caption{Real-world random DiT-2.2B training configuration. FT and RTC steps are stage budgets.}
    \label{tab:supp_real_world_training}
    \small
    \begin{tabular*}{\columnwidth}{@{\extracolsep{\fill}}lr@{}}
        \toprule
        Setting & Value \\
        \midrule
        Demonstrations per task & 100 \\
        FT / RTC stage budget & 30,000 / 15,000 steps \\
        GPU workers per task & 4 \\
        Microbatch per worker & 4 \\
        Gradient accumulation & 2 \\
        Effective batch size & 32 \\
        Precision & BF16 \\
        Optimizer-state partitioning & ZeRO-1 \\
        Optimizer & Fused AdamW \\
        Video-expert initialization seed & 42 \\
        \bottomrule
    \end{tabular*}
\end{table}

The random DiT-2.2B configuration first performs fine-tuning (FT), followed by an RTC stage. RTC starts from the completed FT model with a new optimizer and stage random state. Its configuration uses a maximum prefix delay of 12 and an execution horizon of eight. Resumption within the same stage restores the optimizer, random state, and step count.

\subsection{Real-World Evaluation}

\paragraph{Platform and observations.}
Experiments use an AgileX ALOHA bimanual platform with top, left-wrist, and right-wrist RGB cameras recording at 30 frames per second. Observation preprocessing and action representation are described in Sec.~\ref{sec:supp_vace_inputs}.

\paragraph{Tasks and success criteria.}
The four tasks are Stack Three Bowls, Hang Cup on Mug Tree, Place Wooden Block on Towel, and Place Beverage and Lift Basket. Each policy is evaluated for 20 trials per task, totaling 80 trials. Success requires autonomous task completion within 30 seconds without human intervention; no partial credit is assigned. Overall success pools successful trials across the four tasks.

\section{Detailed Results of TOVS}
\label{sec:supp_current_results}

We report task- and suite-level results for the final policy trained with Training-Only Video Supervision (TOVS). This configuration uses a random DiT-2.2B video expert alongside pretrained $\pi_{0.5}$ during training and deploys the action policy alone.

\subsection{RoboTwin 2.0}

Tables~\ref{tab:final_policy_robotwin_part1} and~\ref{tab:final_policy_robotwin_part2} report all 50 RoboTwin tasks. C2C and C2R denote Clean$\rightarrow$Clean and Clean$\rightarrow$Random, respectively, with 100 evaluation episodes per task and condition. Overall success is 84.84\% (4,242/5,000) for C2C and 57.52\% (2,876/5,000) for C2R, a gap of 27.32 percentage points. The tables describe the final policy's performance distribution and do not contain per-task comparisons with the base policy.

\begin{table}[t]
    \centering
    \caption{TOVS success rates (\%) on RoboTwin, part 1 of 2. Each task uses 100 episodes per condition. C2C: Clean$\rightarrow$Clean; C2R: Clean$\rightarrow$Random.}
    \label{tab:final_policy_robotwin_part1}
    \small
    \setlength{\tabcolsep}{3pt}
    \begin{tabular*}{\columnwidth}{@{\extracolsep{\fill}}lrr@{}}
        \toprule
        Task & C2C & C2R \\
        \midrule
        adjust bottle & 100 & 85 \\
        beat block hammer & 91 & 44 \\
        blocks ranking rgb & 95 & 86 \\
        blocks ranking size & 75 & 60 \\
        click alarmclock & 94 & 60 \\
        click bell & 98 & 69 \\
        dump bin bigbin & 97 & 48 \\
        grab roller & 100 & 92 \\
        handover block & 84 & 32 \\
        handover mic & 99 & 54 \\
        hanging mug & 43 & 20 \\
        lift pot & 100 & 48 \\
        move can pot & 73 & 18 \\
        move pillbottle pad & 96 & 26 \\
        move playingcard away & 97 & 59 \\
        move stapler pad & 37 & 24 \\
        open laptop & 92 & 63 \\
        open microwave & 29 & 12 \\
        pick diverse bottles & 86 & 60 \\
        pick dual bottles & 94 & 65 \\
        place a2b left & 87 & 57 \\
        place a2b right & 81 & 55 \\
        place bread basket & 93 & 76 \\
        place bread skillet & 84 & 68 \\
        place burger fries & 96 & 95 \\
        \bottomrule
    \end{tabular*}
\end{table}

\begin{table}[t]
    \centering
    \caption{TOVS success rates (\%) on RoboTwin, part 2 of 2. Each task uses 100 episodes per condition. C2C: Clean$\rightarrow$Clean; C2R: Clean$\rightarrow$Random. Overall counts are 4,242/5,000 for C2C and 2,876/5,000 for C2R.}
    \label{tab:final_policy_robotwin_part2}
    \small
    \setlength{\tabcolsep}{3pt}
    \begin{tabular*}{\columnwidth}{@{\extracolsep{\fill}}lrr@{}}
        \toprule
        Task & C2C & C2R \\
        \midrule
        place can basket & 78 & 42 \\
        place cans plasticbox & 94 & 87 \\
        place container plate & 99 & 90 \\
        place dual shoes & 98 & 70 \\
        place empty cup & 98 & 77 \\
        place fan & 86 & 53 \\
        place mouse pad & 64 & 31 \\
        place object basket & 91 & 52 \\
        place object scale & 86 & 59 \\
        place object stand & 94 & 77 \\
        place phone stand & 83 & 49 \\
        place shoe & 97 & 79 \\
        press stapler & 97 & 48 \\
        put bottles dustbin & 85 & 24 \\
        put object cabinet & 51 & 26 \\
        rotate qrcode & 84 & 14 \\
        scan object & 83 & 52 \\
        shake bottle & 100 & 97 \\
        shake bottle horizontally & 100 & 98 \\
        stack blocks three & 53 & 43 \\
        stack blocks two & 99 & 95 \\
        stack bowls three & 79 & 69 \\
        stack bowls two & 98 & 79 \\
        stamp seal & 69 & 35 \\
        turn switch & 55 & 54 \\
        \midrule
        Overall & 84.84 & 57.52 \\
        \bottomrule
    \end{tabular*}
\end{table}

Success decreases from C2C to C2R on all 50 tasks, with substantial variation in the size of the decrease. Under randomized scenes, \texttt{shake bottle horizontally} and \texttt{place burger fries} retain 98\% and 95\% success, whereas \texttt{rotate qrcode} changes from 84\% to 14\% and \texttt{move pillbottle pad} from 96\% to 26\%. Six tasks achieve at least 90\% success in C2R, while eight remain below 30\%. Meanwhile, \texttt{open microwave} and \texttt{move stapler pad} reach only 29\% and 37\% in C2C, showing that these tasks are difficult for this policy even in clean scenes.

\subsection{LIBERO and LIBERO Plus}

Table~\ref{tab:final_policy_libero} gives the suite-level rates and counts on LIBERO and LIBERO Plus. On standard LIBERO, TOVS achieves 99.60\%, 100.00\%, 97.00\%, and 95.60\% on Spatial, Object, Goal, and Long, respectively, with 98.05\% overall. On LIBERO Plus, the corresponding rates are 91.17\%, 88.52\%, 81.90\%, and 79.40\%, with 85.15\% overall. Long denotes LIBERO-10.

\begin{table*}[t]
    \centering
    \caption{TOVS results on LIBERO and LIBERO Plus. Each suite reports success rate (\%) and successes/episodes. Long denotes LIBERO-10. Overall rates pool successful episodes across suites.}
    \label{tab:final_policy_libero}
    \small
    \setlength{\tabcolsep}{5pt}
    \begin{tabular*}{\textwidth}{@{\extracolsep{\fill}}lrrrr@{}}
        \toprule
        Suite & \multicolumn{2}{c}{LIBERO} & \multicolumn{2}{c}{LIBERO Plus} \\
        \cmidrule(lr){2-3}\cmidrule(lr){4-5}
        & Rate (\%) & Successes/Episodes & Rate (\%) & Successes/Episodes \\
        \midrule
        Spatial & 99.60 & 498/500 & 91.17 & 2,190/2,402 \\
        Object & 100.00 & 500/500 & 88.52 & 2,229/2,518 \\
        Goal & 97.00 & 485/500 & 81.90 & 2,122/2,591 \\
        Long & 95.60 & 478/500 & 79.40 & 2,000/2,519 \\
        \midrule
        Overall & 98.05 & 1,961/2,000 & 85.15 & 8,541/10,030 \\
        \bottomrule
    \end{tabular*}
\end{table*}

The differences between standard LIBERO and LIBERO Plus are 8.43, 11.48, 15.10, and 16.20 percentage points for Spatial, Object, Goal, and Long, respectively. The larger gaps in Goal and Long describe performance under the two evaluation conditions; the differing scenarios do not isolate a particular perturbation or task horizon as their cause. Because LIBERO Plus has unequal suite budgets, its overall rate pools all 10,030 episodes rather than averaging the four suite percentages.

\subsection{Real-World Results}

The main paper reports real-world evaluation results based on 20 trials per task and policy. Across the four tasks, the base $\pi_{0.5}$ policy succeeds in 34/80 trials (42.50\%), compared with 49/80 (61.25\%) for TOVS trained with random DiT-2.2B. This is an increase of 15 successful trials, or 18.75 percentage points. Equal trial counts make the mean task rate identical to the pooled rate. Per-task success counts are reported in Table~\ref{tab:supp_real_world_counts}.

\begin{table}[t]
    \centering
    \caption{Real-world success counts, with 20 trials per task and policy. TOVS uses random DiT-2.2B during training.}
    \label{tab:supp_real_world_counts}
    \small
    \setlength{\tabcolsep}{3pt}
    \begin{tabular*}{\columnwidth}{@{\extracolsep{\fill}}lcc@{}}
        \toprule
        Task & $\pi_{0.5}$ & TOVS \\
        \midrule
        Stack Three Bowls & 7/20 & 15/20 \\
        Hang Cup on Mug Tree & 2/20 & 3/20 \\
        Wooden Block on Towel & 14/20 & 17/20 \\
        Beverage and Lift Basket & 11/20 & 14/20 \\
        \midrule
        Overall & 34/80 & 49/80 \\
        \bottomrule
    \end{tabular*}
\end{table}

Stack Three Bowls changes from 35\% to 75\%, a gain of 40 percentage points; Hang Cup on Mug Tree changes from 10\% to 15\%, a gain of five points. Wooden Block on Towel changes from 70\% to 85\%, and Beverage and Lift Basket from 55\% to 70\%, both gains of 15 points. Cup hanging remains a low-success task for both policies, illustrating the task variation underlying the aggregate improvement.


\section{Qualitative Execution Comparisons}
\label{sec:supp_qualitative_execution}

\begin{figure*}[!t]
    \centering
    \includegraphics[width=0.98\textwidth,height=0.70\textheight,keepaspectratio]{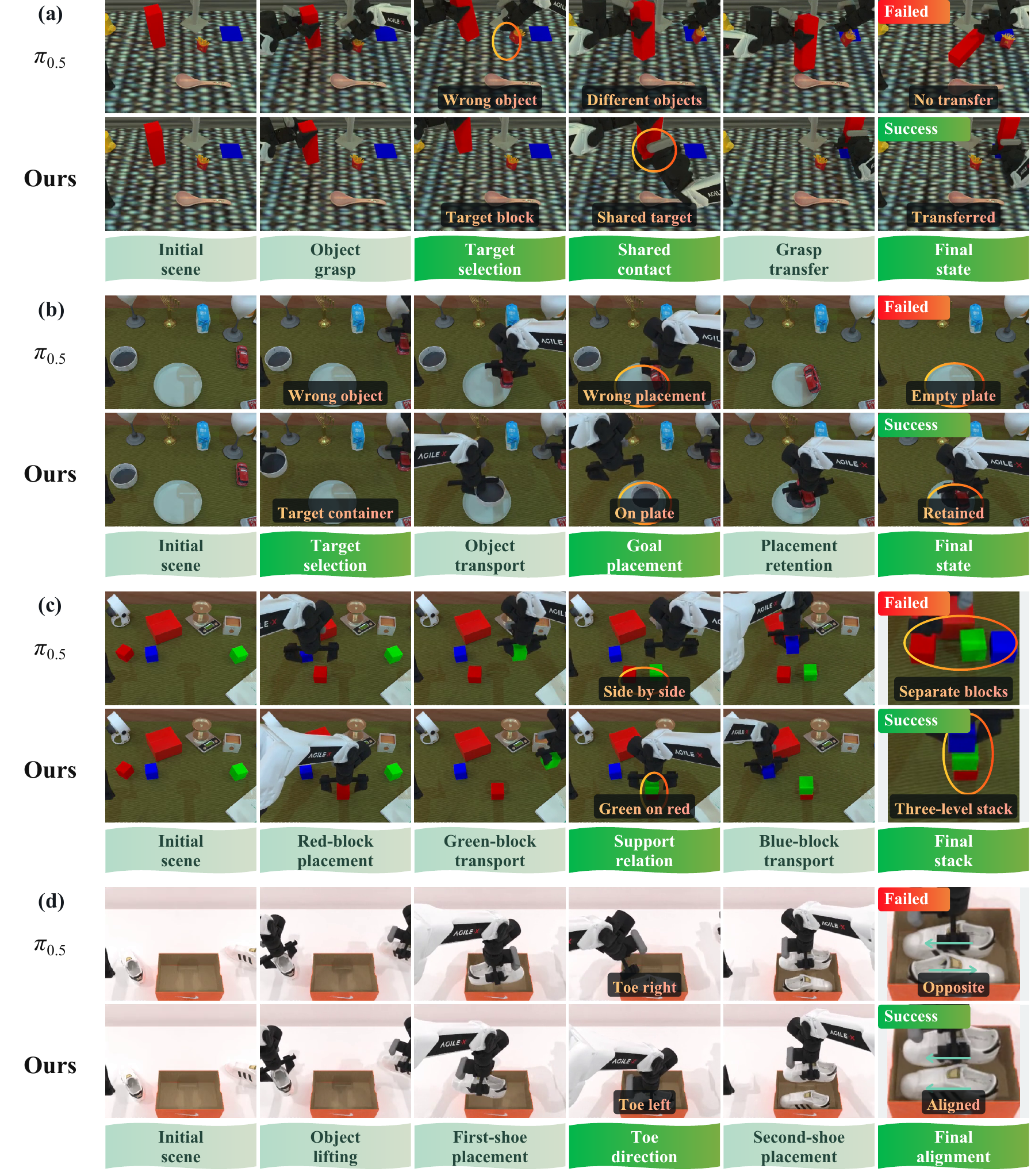}
    \caption{Selected RoboTwin execution comparisons: $\pi_{0.5}$ (top) and TOVS, labeled Ours (bottom), trained with random DiT-2.2B video supervision. (a) Block handover: distractor selection versus target transfer. (b) Container placement: empty plate versus retained container placement. (c) Stacking: separate blocks versus a three-level stack. (d) Shoes: opposite versus matching orientations. The first five columns share source-frame indices; the last shows each rollout's own endpoint. Orange rings mark task-relevant regions, and mint arrows indicate shoe-toe directions. Green strips label analysis stages; darker green highlights selected stages rather than measured attention or action weights.}
    \label{fig:supp_cases}
\end{figure*}

We compare four selected paired RoboTwin rollouts of the base $\pi_{0.5}$ policy and Training-Only Video Supervision (TOVS). ``Ours'' in Fig.~\ref{fig:supp_cases} denotes TOVS trained with a random DiT-2.2B video expert. Block handover, container placement, and three-block stacking use Clean$\rightarrow$Random scenes; dual-shoe placement uses a Clean$\rightarrow$Clean scene. The base policy fails and TOVS succeeds in each selected pair.

In each panel, the first five columns align the policies by source-video frame index, and the last column shows each rollout's own endpoint. Matching spatial crops enlarge the endpoint arrangements for stacking and shoe placement. Frame indices describe recorded progression rather than wall-clock execution time, and equal indices need not correspond to the same action stage.  The scene conditions and seeds, selected source-video frame indices, and rollout endpoints for all four panels are listed in Table~\ref{tab:supp_case_recordings}.

\paragraph{Coordination around the handover target.}
In \texttt{handover\_block} (Fig.~\ref{fig:supp_cases}(a)), both policies initially grasp the red block with the giving arm. The base policy then picks up a fries-shaped distractor with the receiving arm while the giving arm retains the block, leaving the arms manipulating different objects. TOVS brings the receiving gripper to the red block, establishes two-gripper contact, and withdraws the giving arm while retaining the target with the receiving arm. The visible difference occurs in coordinating around the same task object during the transfer of grasp responsibility.

\paragraph{Target selection and preservation of placement.}
In \texttt{place\_container\_plate} (Fig.~\ref{fig:supp_cases}(b)), the base policy first transports a red car-shaped distractor to the plate, leaving the target container at its initial location. Its later container movement does not establish the required container-on-plate arrangement, and the plate is empty at the endpoint. TOVS places the container on the plate first. It subsequently also interacts with the red object, but the container remains on the plate at the successful endpoint. This rollout shows earlier selection of the task container and preservation of its placement, while non-target interactions remain present.

\paragraph{Consecutive support relations.}
Both policies acquire and transport the red, green, and blue blocks in \texttt{stack\_blocks\_three} (Fig.~\ref{fig:supp_cases}(c)). The base policy leaves them side by side on the table. TOVS places green on red and then blue on green, retaining the first support relation while establishing the next. The distinguishing behavior occurs in the goal-relevant placements after successful object acquisition by both policies.

\paragraph{Orientation-sensitive placement.}
Both policies transfer the two shoes into the box in \texttt{place\_dual\_shoes} (Fig.~\ref{fig:supp_cases}(d)). The base policy leaves the shoes pointing in opposite directions, whereas TOVS places them side by side in the same direction. This Clean$\rightarrow$Clean example shows a difference in the final object orientation and arrangement after grasping and transport have succeeded for both policies.

These trajectories provide concrete comparisons of task-object coordination and goal-relevant placement, including failures that arise after successful grasping. They complement the aggregate evaluations, but the selected failure-to-success pairs do not measure the prevalence of these differences or establish a general robustness mechanism.

\begin{table}[t]
    \centering
    \caption{Recording identifiers and selected source-video frames for Fig.~\ref{fig:supp_cases}. C2C and C2R denote Clean$\rightarrow$Clean and Clean$\rightarrow$Random. Endpoints use frame index (recorded rollout steps).}
    \label{tab:supp_case_recordings}
    \scriptsize
    \setlength{\tabcolsep}{2pt}
    \begin{tabular*}{\columnwidth}{@{\extracolsep{\fill}}cll@{}}
        \toprule
        Panel & Scene / seed & Paired frames; endpoints \\
        \midrule
        (a) & \shortstack[l]{C2R\\100002} & \shortstack[l]{0, 54, 84, 134, 224\\Base 799 (800); TOVS 259 (260)} \\
        \addlinespace
        (b) & \shortstack[l]{C2R\\100001} & \shortstack[l]{0, 84, 134, 174, 224\\Base 399 (400); TOVS 227 (228)} \\
        \addlinespace
        (c) & \shortstack[l]{C2R\\100001} & \shortstack[l]{0, 160, 260, 320, 410\\Base 1199 (1200); TOVS 464 (465)} \\
        \addlinespace
        (d) & \shortstack[l]{C2C\\100004} & \shortstack[l]{0, 80, 120, 170, 200\\Base 599 (600); TOVS 228 (229)} \\
        \bottomrule
    \end{tabular*}
\end{table}
\FloatBarrier

\section{Video-Prediction Loss Studies in WAMs}
\label{sec:wam_video_loss_studies}
\label{sec:supp_wam_studies}

\begin{table*}[t]
\centering
\caption{Video-prediction loss ablation success rates (\%). RoboTwin uses 5,000 episodes per configuration and condition; standard LIBERO uses 2,000 episodes per configuration.}
\label{tab:video_loss_ablation}
\small
\setlength{\tabcolsep}{12pt}
\begin{tabular*}{\textwidth}{@{\extracolsep{\fill}}llrrr}
\toprule
Model & Training objective & LIBERO & RoboTwin Clean & RoboTwin Random \\
\midrule
FastWAM  & w/o Video Loss & 94.90 & 31.46 & 5.70 \\
FastWAM  & w/ Video Loss  & 97.30 & 65.14 & 4.86 \\
OpenWAM  & w/o Video Loss & 93.00 & 21.04 & 2.52 \\
OpenWAM  & w/ Video Loss  & 96.65 & 84.58 & 2.18 \\
\bottomrule
\end{tabular*}
\end{table*}

\begin{table*}[t]
\centering
\caption{Standard LIBERO success rates (\%) for the video-prediction loss ablation. Long denotes LIBERO-10.}
\label{tab:libero_video_loss_ablation}
\small
\setlength{\tabcolsep}{7pt}
\begin{tabular*}{\textwidth}{@{\extracolsep{\fill}}llrrrrrr}
\toprule
Model & Training objective & Spatial & Goal & Object & Long & Overall & Successes/Episodes \\
\midrule
FastWAM & w/o Video Loss & 94.80 & 95.40 & 98.00 & 91.40 & 94.90 & 1,898/2,000 \\
FastWAM & w/ Video Loss  & 98.00 & 98.20 & 99.60 & 93.40 & 97.30 & 1,946/2,000 \\
OpenWAM & w/o Video Loss & 96.80 & 94.20 & 96.60 & 84.40 & 93.00 & 1,860/2,000 \\
OpenWAM & w/ Video Loss  & 97.80 & 96.80 & 98.20 & 93.80 & 96.65 & 1,933/2,000 \\
\bottomrule
\end{tabular*}
\end{table*}
\begin{table*}[t]
\centering
\caption{Success rates (\%) under held-out task and environmental distribution shifts. Each model configuration has 1,600 episodes per Seen condition and 3,200 per Unseen condition. $\Delta$ denotes the change from w/o Video Loss to w/ Video Loss in percentage points. Differences are computed before rounding, so displayed rates and differences may differ by 0.01 points when subtracted directly.}
\label{tab:distribution_shift_video_loss}
\small
\setlength{\tabcolsep}{10pt}
\begin{tabular*}{\textwidth}{@{\extracolsep{\fill}}llrrr}
\toprule
Model & Condition & w/o Video Loss & w/ Video Loss & $\Delta$ (pp) \\
\midrule
FastWAM & Seen + Clean        & 18.44 & 54.81 & +36.38 \\
FastWAM & Seen + Randomized   & 0.50  & 0.38  & -0.13 \\
FastWAM & Unseen + Clean      & 15.81 & 16.09 & +0.28 \\
FastWAM & Unseen + Randomized & 3.19  & 1.56  & -1.63 \\
\midrule
OpenWAM & Seen + Clean        & 1.81  & 91.06 & +89.25 \\
OpenWAM & Seen + Randomized   & 0.38  & 0.13  & -0.25 \\
OpenWAM & Unseen + Clean      & 6.38  & 29.75 & +23.38 \\
OpenWAM & Unseen + Randomized & 3.22  & 0.47  & -2.75 \\
\bottomrule
\end{tabular*}
\end{table*}

We additionally examine video-prediction supervision in FastWAM and OpenWAM through loss ablations and evaluations under task and environmental shifts. These studies use independent training and evaluation protocols and characterize the effect of the video loss within each WAM.

\subsection{Video-Prediction Loss Ablation}

\paragraph{Experimental setup.}
For each WAM, we compare training with the video-prediction loss (w/ Video Loss) against training with that loss removed (w/o Video Loss). Both configurations retain the original model architecture, including the video-prediction branch and its associated components. The ablation removes the video-prediction loss term from the training objective; it does not remove the video branch. Within each model, the two configurations use the same model initialization, robot training data, and training budget.

On standard LIBERO, both models are trained for 10 epochs, corresponding to 21,700 steps for FastWAM and 42,730 steps for OpenWAM. Evaluation covers all 40 tasks, with 50 episodes per task and 2,000 episodes per configuration. On RoboTwin 2.0, evaluation covers all 50 tasks, with 100 episodes per task under each of the Clean and Random conditions, yielding 5,000 episodes per condition and 10,000 episodes per configuration. We report the two RoboTwin conditions separately in Table~\ref{tab:video_loss_ablation}.


\paragraph{Standard and clean conditions.}
Adding the video-prediction loss increases the aggregate success rate on standard LIBERO from 94.90\% to 97.30\% for FastWAM and from 93.00\% to 96.65\% for OpenWAM, gains of 2.40 and 3.65 percentage points, respectively. On RoboTwin Clean, FastWAM increases from 31.46\% to 65.14\%, and OpenWAM increases from 21.04\% to 84.58\%, gains of 33.68 and 63.54 percentage points. These results show improved performance on the training tasks under the evaluated standard and clean conditions in both WAM configurations.

Table~\ref{tab:libero_video_loss_ablation} provides the suite-level LIBERO results. Both models improve in each of the Spatial, Goal, Object, and Long suites. The largest suite-level gain is observed for OpenWAM on Long, which increases from 84.40\% to 93.80\%, a gain of 9.40 percentage points. The suite-level results show that the aggregate improvements are not confined to a single suite.

\paragraph{Randomized conditions.}
Video supervision reduces Random success from 5.70\% to 4.86\% for FastWAM and from 2.52\% to 2.18\% for OpenWAM, decreases of 0.84 and 0.34 percentage points. Thus, the improvements under clean evaluation do not extend to the randomized condition in these configurations. Low baseline success limits further interpretation of the behavioral changes responsible for these differences.

\subsection{Held-Out Tasks and Environmental Shifts}

\paragraph{Experimental setup.}
In a separate distribution-shift study, each model is trained on 40 robot manipulation tasks. Evaluation uses five tasks from the training set (Seen) and ten tasks held out from robot training in this study (Unseen). Both groups are evaluated under Clean and Randomized environments. Each task uses 10 fixed initial states, with 32 rollouts per state. Consequently, each model configuration has $5\times10\times32=1{,}600$ episodes per Seen condition and $10\times10\times32=3{,}200$ episodes per Unseen condition. Table~\ref{tab:distribution_shift_video_loss} reports the four conditions separately.

\paragraph{Clean conditions.}
Under Seen + Clean, FastWAM increases from 18.44\% to 54.81\%, and OpenWAM increases from 1.81\% to 91.06\%. On Unseen + Clean, FastWAM increases from 15.81\% to 16.09\% ($+0.28$ percentage points), whereas OpenWAM increases from 6.38\% to 29.75\% ($+23.38$ percentage points). The effect on held-out task success is therefore heterogeneous across these WAM configurations, with a much larger observed gain for OpenWAM.

\paragraph{Randomized conditions.}
All four model--task-group combinations have lower aggregate success with video supervision. On Seen tasks, FastWAM decreases from 0.50\% to 0.38\%, and OpenWAM from 0.38\% to 0.13\%. On Unseen tasks, FastWAM decreases from 3.19\% to 1.56\%, and OpenWAM from 3.22\% to 0.47\%. These observations show no improvement under the environmental shifts evaluated in this study.

\paragraph{Scope of the findings.}
The WAM studies show that improved clean-task execution does not automatically extend to task and environmental shifts. Held-out clean gains vary across configurations, and randomized success decreases in every reported comparison. These findings characterize the scope of video supervision in the evaluated WAMs; they do not isolate the cause of performance differences between WAMs and TOVS.


\section{Limitations and Future Directions}
\label{sec:supp_limitations}

The evidence for Training-Only Video Supervision (TOVS) centers on pretrained $\pi_{0.5}$ within fixed training budgets and the evaluated seed coverage; broader training-seed variability remains unmeasured. Environmental randomization still leaves difficult tasks. The auxiliary video expert and offline feature preparation add training computation, memory, and storage costs, even though the expert is removed at deployment. Real-world evaluation uses 20 trials per task and policy, limiting the precision and coverage of the estimates.

\paragraph{Other VLA backbones.}
Evaluating TOVS with other pretrained VLA backbones under matched data, training budgets, and evaluation conditions would clarify its applicability beyond the policy studied here.

\paragraph{Expert capacity, initialization, and robot data.}
Controlled studies that separately vary video-expert capacity, initialization, and robot-data scale could distinguish their effects. Within-architecture initialization comparisons and cross-architecture expert comparisons should be analyzed separately, with the policy backbone, budget, and loss settings matched where feasible. Such studies could also test whether the starting policy's capabilities affect the benefit of auxiliary supervision.

\paragraph{Critical execution stages.}
Future work could combine video prediction with auxiliary targets for contact states, failure recovery, and task completion. Stage-level annotations and targeted ablations would help identify benefits during target selection, bimanual handover, and successive relational placements, and test whether stage-focused supervision improves performance or training efficiency.

\end{document}